\documentclass[conference,a4paper]{IEEEtran}
\IEEEoverridecommandlockouts

\usepackage{booktabs}
\usepackage{tabularx}

\usepackage{tikz}

\usepackage{cite}
\usepackage{amsmath,amssymb,amsfonts}
\usepackage{algorithmic}
\usepackage{graphicx}
\usepackage{textcomp}
\usepackage{xcolor}

\usepackage{url}
\usepackage{hyperref}
\usepackage{soul}

\usepackage[acronyms,nonumberlist,nopostdot,nomain,nogroupskip,acronymlists={hidden}]{glossaries}
\newglossary[algh]{hidden}{acrh}{acnh}{Hidden Acronyms}
\glsdisablehyper
\newacronym{3gpp}{3GPP}{3rd Generation Partnership Project}
\newacronym{4g}{4G}{4th generation mobile network}
\newacronym{5g}{5G}{5th generation mobile network}
\newacronym{6g}{6G}{6th generation mobile network}
\newacronym{nextg}{NextG}{Next Generation}
\newacronym{5gc}{5GC}{5G Core}
\newacronym{adc}{ADC}{Analog to Digital Converter}
\newacronym{aerpaw}{AERPAW}{Aerial Experimentation and Research Platform for Advanced Wireless}
\newacronym{ai}{AI}{Artificial Intelligence}
\newacronym{aimd}{AIMD}{Additive Increase Multiplicative Decrease}
\newacronym{am}{AM}{Acknowledged Mode}
\newacronym{amc}{AMC}{Adaptive Modulation and Coding}
\newacronym{amf}{AMF}{Access and Mobility Management Function}
\newacronym{aops}{AOPS}{Adaptive Order Prediction Scheduling}
\newacronym{api}{API}{Application Programming Interface}
\newacronym{apn}{APN}{Access Point Name}
\newacronym{aqm}{AQM}{Active Queue Management}
\newacronym{ausf}{AUSF}{Authentication Server Function}
\newacronym{avc}{AVC}{Advanced Video Coding}
\newacronym{awgn}{AGWN}{Additive White Gaussian Noise}
\newacronym{balia}{BALIA}{Balanced Link Adaptation Algorithm}
\newacronym{bbu}{BBU}{Base Band Unit}
\newacronym{bdp}{BDP}{Bandwidth-Delay Product}
\newacronym{ber}{BER}{Bit Error Rate}
\newacronym{bf}{BF}{Beamforming}
\newacronym{bler}{BLER}{Block Error Rate}
\newacronym{brr}{BRR}{Bayesian Ridge Regressor}
\newacronym{bsr}{BSR}{Buffer Status Report}
\newacronym{bs}{BS}{Base Station}
\newacronym{bpsk}{BPSK}{Binary Phase-shift keying}
\newacronym{bss}{BSS}{Business Support System}
\newacronym{ca}{CA}{Carrier Aggregation}
\newacronym{caas}{CaaS}{Connectivity-as-a-Service}
\newacronym{cb}{CB}{Code Block}
\newacronym{cc}{CC}{Congestion Control}
\newacronym{ccid}{CCID}{Congestion Control ID}
\newacronym{cco}{CC}{Carrier Component}
\newacronym{cd}{CD}{Continuous Delivery}
\newacronym{cdd}{CDD}{Cyclic Delay Diversity}
\newacronym{cdf}{CDF}{Cumulative Distribution Function}
\newacronym{cdma}{CDMA}{Code-Division Multiple Access}
\newacronym{cdn}{CDN}{Content Distribution Network}
\newacronym{ci}{CI}{Continuous Integration}
\newacronym{cicd}{CI/CD}{Continuous Integration/Continuous Delivery}
\newacronym{cir}{CIR}{Channel Impulse Response}
\newacronym{cn}{CN}{Core Network}
\newacronym{codel}{CoDel}{Controlled Delay Management}
\newacronym{comac}{COMAC}{Converged Multi-Access and Core}
\newacronym{cord}{CORD}{Central Office Re-architected as a Datacenter}
\newacronym{cornet}{CORNET}{COgnitive Radio NETwork}
\newacronym{cosmos}{COSMOS}{Cloud Enhanced Open Software Defined Mobile Wireless Testbed for City-Scale Deployment}
\newacronym{cots}{COTS}{Commercial Off-the-Shelf}
\newacronym{cp}{CP}{Control Plane}
\newacronym{cpu}{CPU}{Central Processing Unit}
\newacronym{cqi}{CQI}{Channel Quality Information}
\newacronym{cr}{CR}{Cognitive Radio}
\newacronym{cran}{CRAN}{Cloud \gls{ran}}
\newacronym{crs}{CRS}{Cell Reference Signal}
\newacronym{csi}{CSI}{Channel State Information}
\newacronym{csirs}{CSI-RS}{Channel State Information - Reference Signal}
\newacronym{cu}{CU}{Central Unit}
\newacronym{d2tcp}{D$^2$TCP}{Deadline-aware Data center TCP}
\newacronym{d3}{D$^3$}{Deadline-Driven Delivery}
\newacronym{dac}{DAC}{Digital to Analog Converter}
\newacronym{dag}{DAG}{Directed Acyclic Graph}
\newacronym{darpa}{DARPA}{Defense Advanced Research Projects Agency}
\newacronym{das}{DAS}{Distributed Antenna System}
\newacronym{dash}{DASH}{Dynamic Adaptive Streaming over HTTP}
\newacronym{dc}{DC}{Dual Connectivity}
\newacronym{dccp}{DCCP}{Datagram Congestion Control Protocol}
\newacronym{dce}{DCE}{Direct Code Execution}
\newacronym{dci}{DCI}{Downlink Control Information}
\newacronym{dcl}{DCL}{Dear Colleague Letter}
\newacronym{dctcp}{DCTCP}{Data Center TCP}
\newacronym{devops}{DevOps}{Development and Operations}
\newacronym{dl}{DL}{Deep Learning}
\newacronym{dmr}{DMR}{Deadline Miss Ratio}
\newacronym{dmrs}{DMRS}{DeModulation Reference Signal}
\newacronym{drlcc}{DRL-CC}{Deep Reinforcement Learning Congestion Control}
\newacronym{drs}{DRS}{Discovery Reference Signal}
\newacronym{dt}{DT}{Digital Twin}
\newacronym{dtn}{DTN}{Digital Twin Network}
\newacronym{dtmn}{DTMN}{Digital Twin for Mobile Network}
\newacronym{dtwn}{DTWN}{Digital Twin Wireless Network}
\newacronym{du}{DU}{Distributed Unit}
\newacronym{e2e}{E2E}{end-to-end}
\newacronym{ecaas}{ECaaS}{Edge-Cloud-as-a-Service}
\newacronym{ecn}{ECN}{Explicit Congestion Notification}
\newacronym{edf}{EDF}{Earliest Deadline First}
\newacronym{em}{EM}{Electro-Magnetic}
\newacronym{embb}{eMBB}{Enhanced Mobile Broadband}
\newacronym{empower}{EMPOWER}{EMpowering transatlantic PlatfOrms for advanced WirEless Research}
\newacronym{enb}{eNB}{evolved Node Base}
\newacronym{endc}{EN-DC}{E-UTRAN-\gls{nr} \gls{dc}}
\newacronym{epc}{EPC}{Evolved Packet Core}
\newacronym{eps}{EPS}{Evolved Packet System}
\newacronym{es}{ES}{Edge Server}
\newacronym{etsi}{ETSI}{European Telecommunications Standards Institute}
\newacronym[firstplural=Estimated Times of Arrival (ETAs)]{eta}{ETA}{Estimated Time of Arrival}
\newacronym{eutran}{E-UTRAN}{Evolved Universal Terrestrial Access Network}
\newacronym{faas}{FaaS}{Function-as-a-Service}
\newacronym{fapi}{FAPI}{Functional Application Platform Interface}
\newacronym{fcc}{FCC}{Federal Communications Commission}
\newacronym{fdd}{FDD}{Frequency Division Duplexing}
\newacronym{fdm}{FDM}{Frequency Division Multiplexing}
\newacronym{fdma}{FDMA}{Frequency Division Multiple Access}
\newacronym{fed4fire}{FED4FIRE+}{Federation 4 Future Internet Research and Experimentation Plus}
\newacronym{fir}{FIR}{Finite Impulse Response}
\newacronym{fit}{FIT}{Future \acrlong{iot}}
\newacronym{fl}{FL}{Federated Learning}
\newacronym{fpga}{FPGA}{Field Programmable Gate Array}
\newacronym{fr2}{FR2}{Frequency Range 2}
\newacronym{fs}{FS}{Fast Switching}
\newacronym{fscc}{FSCC}{Flow Sharing Congestion Control}
\newacronym{ftp}{FTP}{File Transfer Protocol}
\newacronym{fw}{FW}{Flow Window}
\newacronym{ga128}{Ga}{Golay Sequence type A}
\newacronym{ge}{GE}{Gaussian Elimination}
\newacronym{glfsr}{GLFSR}{Galois Linear Feedback Shift Register}
\newacronym{gnb}{gNB}{Next Generation Node Base}
\newacronym{gold}{Gold}{Gold}
\newacronym{gop}{GOP}{Group of Pictures}
\newacronym{gpr}{GPR}{Gaussian Process Regressor}
\newacronym{gpu}{GPU}{Graphics Processing Unit}
\newacronym{gtp}{GTP}{GPRS Tunneling Protocol}
\newacronym{gtpc}{GTP-C}{GPRS Tunnelling Protocol Control Plane}
\newacronym{gtpu}{GTP-U}{GPRS Tunnelling Protocol User Plane}
\newacronym{gtpv2c}{GTPv2-C}{\gls{gtp} v2 - Control}
\newacronym{gw}{GW}{Gateway}
\newacronym{harq}{HARQ}{Hybrid Automatic Repeat reQuest}
\newacronym{hetnet}{HetNet}{Heterogeneous Network}
\newacronym{hh}{HH}{Hard Handover}
\newacronym{hol}{HOL}{Head-of-Line}
\newacronym{hqf}{HQF}{Highest-quality-first}
\newacronym{hss}{HSS}{Home Subscription Server}
\newacronym{http}{HTTP}{HyperText Transfer Protocol}
\newacronym{ia}{IA}{Initial Access}
\newacronym{iab}{IAB}{Integrated Access and Backhaul}
\newacronym{ic}{IC}{Incident Command}
\newacronym{ietf}{IETF}{Internet Engineering Task Force}
\newacronym{ifw}{IFW}{Interference Free Window}
\newacronym{imsi}{IMSI}{International Mobile Subscriber Identity}
\newacronym{imt}{IMT}{International Mobile Telecommunication}
\newacronym{iot}{IoT}{Internet of Things}
\newacronym{ip}{IP}{Internet Protocol}
\newacronym{iq}{IQ}{In-phase and Quadrature}
\newacronym{isi}{ISI}{Inter-Symbol Interference}
\newacronym{itu}{ITU}{International Telecommunication Union}
\newacronym{kpi}{KPI}{Key Performance Indicator}
\newacronym{kvm}{KVM}{Kernel-based Virtual Machine}
\newacronym{lfsr}{LFSR}{Linear Feedback Shift Register}
\newacronym{los}{LOS}{Line-of-Sight}
\newacronym{ls}{LS}{Loosely Synchronised}
\newacronym{lsm}{LSM}{Link-to-System Mapping}
\newacronym{lstm}{LSTM}{Long Short Term Memory}
\newacronym{lte}{LTE}{Long Term Evolution}
\newacronym{lxc}{LXC}{Linux Container}
\newacronym{m2m}{M2M}{Machine to Machine}
\newacronym{mac}{MAC}{Medium Access Control}
\newacronym{mai}{MAI}{Multiple Access Interference}
\newacronym{manet}{MANET}{Mobile Ad Hoc Network}
\newacronym{mano}{MANO}{Management and Orchestration}
\newacronym{mc}{MC}{Multi-Connectivity}
\newacronym{mcc}{MCC}{Mobile Cloud Computing}
\newacronym{mchem}{MCHEM}{Massive Channel Emulator}
\newacronym{mcs}{MCS}{Modulation and Coding Scheme}
\newacronym{mec}{MEC}{Multi-access Edge Computing}
\newacronym{mec2}{MEC}{Mobile Edge Cloud}
\newacronym{mec3}{MEC}{Mobile Edge Computing}
\newacronym{mfc}{MFC}{Mobile Fog Computing}
\newacronym{mi}{MI}{Mutual Information}
\newacronym{mib}{MIB}{Master Information Block}
\newacronym{miesm}{MIESM}{Mutual Information Based Effective SINR}
\newacronym{mimo}{MIMO}{Multiple Input, Multiple Output}
\newacronym{mgen}{MGEN}{Multi-Generator}
\newacronym{ml}{ML}{Machine Learning}
\newacronym{mlr}{MLR}{Maximum-local-rate}
\newacronym[plural=\gls{mme}s,firstplural=Mobility Management Entities (MMEs)]{mme}{MME}{Mobility Management Entity}
\newacronym{mmtc}{mMTC}{Massive Machine-Type Communications}
\newacronym{mmwave}{mmWave}{millimeter wave}
\newacronym{mpdccp}{MP-DCCP}{Multipath Datagram Congestion Control Protocol}
\newacronym{mptcp}{MPTCP}{Multipath TCP}
\newacronym{mr}{MR}{Maximum Rate}
\newacronym{mrdc}{MR-DC}{Multi \gls{rat} \gls{dc}}
\newacronym{mse}{MSE}{Mean Square Error}
\newacronym{mss}{MSS}{Maximum Segment Size}
\newacronym{mt}{MT}{Mobile Termination}
\newacronym{mtd}{MTD}{Machine-Type Device}
\newacronym{mtu}{MTU}{Maximum Transmission Unit}
\newacronym{mumimo}{MU-MIMO}{Multi-user \gls{mimo}}
\newacronym{mvno}{MVNO}{Mobile Virtual Network Operator}
\newacronym{nalu}{NALU}{Network Abstraction Layer Unit}
\newacronym{nas}{NAS}{Network Attached Storage}
\newacronym{nbiot}{NB-IoT}{Narrow Band IoT}
\newacronym{nfv}{NFV}{Network Function Virtualization}
\newacronym{nfvi}{NFVI}{Network Function Virtualization Infrastructure}
\newacronym{nic}{NIC}{Network Interface Card}
\newacronym{nlos}{NLOS}{Non-Line-of-Sight}
\newacronym{now}{NOW}{Non Overlapping Window}
\newacronym{nrdz}{NRDZ}{National Radio Dynamic Zone}
\newacronym{nsf}{NSF}{National Science Foundation}
\newacronym{nsm}{NSM}{Network Service Mesh}
\newacronym[type=hidden]{nr}{NR}{New Radio}
\newacronym{nrf}{NRF}{Network Repository Function}
\newacronym{nsa}{NSA}{Non Stand Alone}
\newacronym{nse}{NSE}{Network Slicing Engine}
\newacronym{nssf}{NSSF}{Network Slice Selection Function}
\newacronym{ntp}{NTP}{Network Time Protocol}
\newacronym{o2i}{O2I}{Outdoor to Indoor}
\newacronym{oai}{OAI}{OpenAirInterface}
\newacronym{oaicn}{OAI-CN}{\gls{oai} \acrlong{cn}}
\newacronym{oairan}{OAI-RAN}{\acrlong{oai} \acrlong{ran}}
\newacronym{oam}{OAM}{Operations, Administration and Maintenance}
\newacronym[plural=\gls{obu}s,firstplural=Onboard Units (OBUs)]{obu}{OBU}{Onboard Unit}
\newacronym{ofdm}{OFDM}{Orthogonal Frequency Division Multiplexing}
\newacronym{olia}{OLIA}{Opportunistic Linked Increase Algorithm}
\newacronym{omec}{OMEC}{Open Mobile Evolved Core}
\newacronym{onap}{ONAP}{Open Network Automation Platform}
\newacronym{onf}{ONF}{Open Networking Foundation}
\newacronym{onos}{ONOS}{Open Networking Operating System}
\newacronym{oom}{OOM}{\gls{onap} Operations Manager}
\newacronym{opnfv}{OPNFV}{Open Platform for \gls{nfv}}
\newacronym[type=hidden]{oran}{O-RAN}{Open \gls{ran}}
\newacronym{orbit}{ORBIT}{Open-Access Research Testbed for Next-Generation Wireless Networks}
\newacronym{os}{OS}{Operating System}
\newacronym{osm}{OSM}{Open Street Map}
\newacronym{oss}{OSS}{Operations Support System}
\newacronym{pa}{PA}{Position-aware}
\newacronym{pase}{PASE}{Prioritization, Arbitration, and Self-adjusting Endpoints}
\newacronym{pawr}{PAWR}{Platforms for Advanced Wireless Research}
\newacronym{pbch}{PBCH}{Physical Broadcast Channel}
\newacronym{pcef}{PCEF}{Policy and Charging Enforcement Function}
\newacronym{pcfich}{PCFICH}{Physical Control Format Indicator Channel}
\newacronym{pcrf}{PCRF}{Policy and Charging Rules Function}
\newacronym{pdcch}{PDCCH}{Physical Downlink Control Channel}
\newacronym{pdcp}{PDCP}{Packet Data Convergence Protocol}
\newacronym{pdsch}{PDSCH}{Physical Downlink Shared Channel}
\newacronym{pdu}{PDU}{Packet Data Unit}
\newacronym{pdp}{PDP}{Power Delay Profile}
\newacronym{pf}{PF}{Proportional Fair}
\newacronym{pgw}{PGW}{Packet Gateway}
\newacronym{phich}{PHICH}{Physical Hybrid ARQ Indicator Channel}
\newacronym{phy}{PHY}{Physical}
\newacronym{pl}{PL}{Path Loss}
\newacronym{pmch}{PMCH}{Physical Multicast Channel}
\newacronym{pmi}{PMI}{Precoding Matrix Indicators}
\newacronym{powder}{POWDER}{Platform for Open Wireless Data-driven Experimental Research}
\newacronym{ppo}{PPO}{Proximal Policy Optimization}
\newacronym{ppp}{PPP}{Poisson Point Process}
\newacronym{prach}{PRACH}{Physical Random Access Channel}
\newacronym{prb}{PRB}{Physical Resource Block}
\newacronym{psnr}{PSNR}{Peak Signal to Noise Ratio}
\newacronym{pss}{PSS}{Primary Synchronization Signal}
\newacronym{pucch}{PUCCH}{Physical Uplink Control Channel}
\newacronym{pusch}{PUSCH}{Physical Uplink Shared Channel}
\newacronym{qam}{QAM}{Quadrature Amplitude Modulation}
\newacronym{qci}{QCI}{\gls{qos} Class Identifier}
\newacronym{qoe}{QoE}{Quality of Experience}
\newacronym{qos}{QoS}{Quality of Service}
\newacronym{qtgui}{QT-GUI}{QT Graphical User Interface}
\newacronym{quic}{QUIC}{Quick UDP Internet Connections}
\newacronym{rach}{RACH}{Random Access Channel}
\newacronym{ran}{RAN}{Radio Access Network}
\newacronym[firstplural=Radio Access Technologies (RATs)]{rat}{RAT}{Radio Access Technology}
\newacronym{rcn}{RCN}{Research Coordination Network}
\newacronym{rec}{REC}{Radio Edge Cloud}
\newacronym{red}{RED}{Random Early Detection}
\newacronym{renew}{RENEW}{Reconfigurable Eco-system for Next-generation End-to-end Wireless}
\newacronym{rf}{RF}{Radio Frequency}
\newacronym{rfc}{RFC}{Request for Comments}
\newacronym{rfr}{RFR}{Random Forest Regressor}
\newacronym{ric}{RIC}{\gls{ran} Intelligent Controller}
\newacronym{rlc}{RLC}{Radio Link Control}
\newacronym{rlf}{RLF}{Radio Link Failure}
\newacronym{rlnc}{RLNC}{Random Linear Network Coding}
\newacronym{rmse}{RMSE}{Root Mean Squared Error}
\newacronym{rnis}{RNIS}{Radio Network Information Service}
\newacronym{rr}{RR}{Round Robin}
\newacronym{rrc}{RRC}{Radio Resource Control}
\newacronym{rrm}{RRM}{Radio Resource Management}
\newacronym{rru}{RRU}{Remote Radio Unit}
\newacronym{rs}{RS}{Remote Server}
\newacronym{rsrp}{RSRP}{Reference Signal Received Power}
\newacronym{rsrq}{RSRQ}{Reference Signal Received Quality}
\newacronym{rss}{RSS}{Received Signal Strength}
\newacronym{rssi}{RSSI}{Received Signal Strength Indicator}
\newacronym{rsu}{RSU}{Road-Side Unit}
\newacronym{rtt}{RTT}{Round Trip Time}
\newacronym{ru}{RU}{Radio Unit}
\newacronym{rw}{RW}{Receive Window}
\newacronym{rx}{RX}{Receiver}
\newacronym{s1ap}{S1AP}{S1 Application Protocol}
\newacronym{sa}{SA}{standalone}
\newacronym{sack}{SACK}{Selective Acknowledgment}
\newacronym{sap}{SAP}{Service Access Point}
\newacronym{sc2}{SC2}{Spectrum Collaboration Challenge}
\newacronym{scef}{SCEF}{Service Capability Exposure Function}
\newacronym{sch}{SCH}{Secondary Cell Handover}
\newacronym{scoot}{SCOOT}{Split Cycle Offset Optimization Technique}
\newacronym{sctp}{SCTP}{Stream Control Transmission Protocol}
\newacronym{sdap}{SDAP}{Service Data Adaptation Protocol}
\newacronym{sd}{SD}{Standard Deviation}
\newacronym{sdk}{SDK}{Software Development Kit}
\newacronym{sdm}{SDM}{Space Division Multiplexing}
\newacronym{sdma}{SDMA}{Spatial Division Multiple Access}
\newacronym{sdn}{SDN}{Software-defined Networking}
\newacronym{sdr}{SDR}{Software-defined Radio}
\newacronym{seba}{SEBA}{SDN-Enabled Broadband Access}
\newacronym{sgsn}{SGSN}{Serving GPRS Support Node}
\newacronym{sgw}{SGW}{Service Gateway}
\newacronym{si}{SI}{Study Item}
\newacronym{sib}{SIB}{Secondary Information Block}
\newacronym{sinr}{SINR}{Signal to Interference plus Noise Ratio}
\newacronym{sip}{SIP}{Session Initiation Protocol}
\newacronym{siso}{SISO}{Single Input, Single Output}
\newacronym{sla}{SLA}{Service Level Agreement}
\newacronym{sm}{SM}{Saturation Mode}
\newacronym{smf}{SMF}{Session Management Function}
\newacronym{smo}{SMO}{Service Management and Orchestration}
\newacronym{sms}{SMS}{Short Message Service}
\newacronym{smsgmsc}{SMS-GMSC}{\gls{sms}-Gateway}
\newacronym{snr}{SNR}{Signal-to-Noise-Ratio}
\newacronym{son}{SON}{Self-Organizing Network}
\newacronym{sptcp}{SPTCP}{Single Path TCP}
\newacronym{srb}{SRB}{Service Radio Bearer}
\newacronym{srn}{SRN}{Standard Radio Node}
\newacronym{srs}{SRS}{Sounding Reference Signal}
\newacronym{ss}{SS}{Synchronization Signal}
\newacronym{sss}{SSS}{Secondary Synchronization Signal}
\newacronym{st}{ST}{Spanning Tree}
\newacronym{svc}{SVC}{Scalable Video Coding}
\newacronym{tb}{TB}{Transport Block}
\newacronym{tcp}{TCP}{Transmission Control Protocol}
\newacronym{tdd}{TDD}{Time Division Duplexing}
\newacronym{tdm}{TDM}{Time Division Multiplexing}
\newacronym{tdma}{TDMA}{Time Division Multiple Access}
\newacronym{tfl}{TfL}{Transport for London}
\newacronym{tfrc}{TFRC}{TCP-Friendly Rate Control}
\newacronym{tft}{TFT}{Traffic Flow Template}
\newacronym{tgen}{TGEN}{Traffic Generator}
\newacronym{tip}{TIP}{Telecom Infra Project}
\newacronym{tm}{TM}{Transparent Mode}
\newacronym{to}{TO}{Telco Operator}
\newacronym{toa}{ToA}{Time of Arrival}
\newacronym{tr}{TR}{Technical Report}
\newacronym{trp}{TRP}{Transmitter Receiver Pair}
\newacronym{ts}{TS}{Technical Specification}
\newacronym{tti}{TTI}{Transmission Time Interval}
\newacronym{ttt}{TTT}{Time-to-Trigger}
\newacronym{tx}{TX}{Transmitter}
\newacronym{uas}{UAS}{Unmanned Aerial System}
\newacronym{uav}{UAV}{Unmanned Aerial Vehicle}
\newacronym{udm}{UDM}{Unified Data Management}
\newacronym{udp}{UDP}{User Datagram Protocol}
\newacronym{udr}{UDR}{Unified Data Repository}
\newacronym{ue}{UE}{User Equipment}
\newacronym{uhd}{UHD}{\gls{usrp} Hardware Driver}
\newacronym{ul}{UL}{Uplink}
\newacronym{um}{UM}{Unacknowledged Mode}
\newacronym{uml}{UML}{Unified Modeling Language}
\newacronym{upa}{UPA}{Uniform Planar Array}
\newacronym{upf}{UPF}{User Plane Function}
\newacronym{urllc}{URLLC}{Ultra Reliable and Low Latency Communication}
\newacronym{usa}{U.S.}{United States}
\newacronym{usim}{USIM}{Universal Subscriber Identity Module}
\newacronym{usrp}{USRP}{Universal Software Radio Peripheral}
\newacronym{utc}{UTC}{Urban Traffic Control}
\newacronym{vim}{VIM}{Virtualization Infrastructure Manager}
\newacronym{vm}{VM}{Virtual Machine}
\newacronym{vnf}{VNF}{Virtual Network Function}
\newacronym{volte}{VoLTE}{Voice over \gls{lte}}
\newacronym{voltha}{VOLTHA}{Virtual OLT HArdware Abstraction}
\newacronym{vr}{VR}{Virtual Reality}
\newacronym{vran}{vRAN}{Virtualized \gls{ran}}
\newacronym{vss}{VSS}{Video Streaming Server}
\newacronym{wbf}{WBF}{Wired Bias Function}
\newacronym{wf}{WF}{Wired-first}
\newacronym{wi}{WI}{Wireless InSite}
\newacronym{wlan}{WLAN}{Wireless Local Area Network}
\newacronym{pnf}{PNF}{Physical Network Function}
\newacronym{drl}{DRL}{Deep Reinforcement Learning}
\newacronym{mtc}{MTC}{Machine-type Communications}
\newacronym{v2x}{V2X}{Vehicle-to-everything}
\newacronym{cast}{CaST}{Channel emulation scenario generator and Sounder Toolchain}
\newacronym{gui}{GUI}{Graphical User Interface}
\newacronym{ups}{UPS}{Uninterruptible Power Supply}
\newacronym{ota}{OTA}{Over-the-Air}
\newacronym{hitl}{HITL}{hardware-in-the-loop}
\newacronym{soc}{SoC}{System-on-Chip}
\newacronym{eeg}{EEG}{electroencephalogram}
\newacronym{ieeg}{iEEG}{intracranial electroencephalogram}
\newacronym{ecg}{ECG}{electrocardiogram}
\newacronym{fph}{FPH}{false positive per hour}
\newacronym{cnn}{CNN}{Convolutional Neural Network}
\newacronym{ban}{BAN}{Body Area Network}
\newacronym{roc}{ROC}{Receiver Operating Characteristic Curve}
\newacronym{dbs}{DBS}{Deep Brain Stimulator}
\newacronym{auc}{AUC}{Area Under the Curve}

\usepackage{etoolbox}
\makeatletter
\patchcmd{\@makecaption}
  {\scshape}
  {}
  {}
  {}
\makeatletter
\patchcmd{\@makecaption}
  {\\}
  {.\ }
  {}
  {}
\makeatother

\def\BibTeX{{\rm B\kern-.05em{\sc i\kern-.025em b}\kern-.08em
    T\kern-.1667em\lower.7ex\hbox{E}\kern-.125emX}}

\begin{document}

\title{SeizNet: An AI-enabled Implantable Sensor Network System for Seizure Prediction

\thanks{This work was partially supported by the U.S.\ National Science Foundation under grant TI-2214013.}
}

\author{\IEEEauthorblockN{Ali Saeizadeh$^\dagger$, Douglas Schonholtz$^\dagger$, Daniel Uvaydov$^\dagger$, Raffaele Guida$^\dagger$, Emrecan Demirors$^\dagger$, \\Pedram Johari$^\dagger$, Jorge M. Jimenez$^\dagger$, Joseph S. Neimat$^*$, Tommaso Melodia$^\dagger$}
\IEEEauthorblockA{$^\dagger$Institute for the Wireless Internet of Things, Northeastern University, Boston, MA, U.S.A.\\
$^*$University of Louisville, Louisville, KY, U.S.A.\\
E-mail: $^\dagger$\{saeizadeh.a, schonholtz.d, uvaydov.d, guida.r, e.demirors,\\
p.johari, j.jimenez, melodia\}@northeastern.edu, $^*$joseph.neimat@uoflhealth.org}
}

\maketitle



\begin{abstract}
In this paper, we introduce SeizNet, a closed-loop system for predicting epileptic seizures through the use of \gls{dl} method and implantable sensor networks.
While pharmacological treatment is effective for some epilepsy patients (with $\sim$65M people affected worldwide), one out of three suffer from drug-resistant epilepsy. To alleviate the impact of seizure, predictive systems have been developed that can notify such patients of an impending seizure, allowing them to take precautionary measures. SeizNet leverages \gls{dl} techniques and combines data from multiple recordings, specifically \gls{ieeg} and \gls{ecg} sensors, that can significantly improve the specificity of seizure prediction while preserving very high levels of sensitivity. SeizNet \gls{dl} algorithms are designed for efficient real-time execution at the edge, minimizing data privacy concerns, data transmission overhead, and power inefficiencies associated with cloud-based solutions. Our results indicate that SeizNet outperforms traditional single-modality and non-personalized prediction systems in all metrics, achieving up to 99\% accuracy in predicting seizure, offering a promising new avenue in refractory epilepsy treatment.
\end{abstract}

\section{Introduction}\label{sec:intro}


Epilepsy is a common neurological disorder disease with around 65M people diagnosed worldwide and a risk of premature death three times higher than that of the general population~\cite{who}. 
Although most patients diagnosed with epilepsy respond well to pharmaceutical drugs to treat epilepsy, approximately one-third of them suffer from drug resistant epilepsy ~\cite{xue2019risk}. 
Therefore, there is a need for alternative epilepsy treatments that goes beyond the pharmaceutical care. Recently, studies has been devoted to predicting seizure onsets well ahead of time in order to notify patients in advance to prevent detrimental accidents with their precautionary actions.
\cite{mormann2007seizure,ramgopal2014seizure, uvaydov2022aieeg}.

One of the main challenges in data-driven seizure prediction techniques arises due to the infrequency of seizures in patient recordings, making these methods prone to biases. Prior research has addressed this issue through under-sampling the non-seizure periods. However, any misjudgment can potentially lead to excessive false positives, i.e., falsely alerting the patients of an upcoming seizure that never happens. 
Furthermore, among many works that use \gls{dl} techniques to predict seizures, 
almost all rely on using a single biological time-series signal such as \gls{eeg}, \gls{ieeg}, or \gls{ecg}, 
with the majority using \gls{eeg} or \gls{ieeg} \cite{kuhlmann2018seizure, park2011seizure, billeci2018patient}. While these single modality prediction techniques have shown to be effective, there is untapped potential in possibly using a combination of different modalities to create prediction with less variance~\cite{schulze2022seizure}. 


In this work we propose an end-to-end framework for multi-modal seizure prediction using \gls{ieeg} and \gls{ecg}, utilizing a sensor network to enhance seizure prediction accuracy at the edge. We utilize an ultrasonic intra-body communication system to facilitate safe, secure and low-power communication between the sensors. We design \gls{dl} structures using both forms of sensor recordings, as well as an effective way for combining the classifications results driven from each sensor's \gls{dl} model classifying pre-seizure (\textit{preictal}) from non-seizure (\textit{interictal}) periods. 
Our framework can achieve the utmost accuracy in seizure prediction, surpassing 99\% in both sensitivity and specificity. We further employ a focal loss function to address the imbalances in \gls{dl} dataset, and showcase the potential of using only \gls{ecg} signals as a non-invasive and easily accessible input for seizure prediction with unprecedented accuracy (up to 94\% sensitivity and 99\% specificity).




In Sec.~\ref{sec:sensor-network} we outline the ultrasonic sensor network model; Sec.~\ref{sec:dataset} explains the dataset; Sec.~\ref{sec:dl-system} presents the proposed combined sensors seizure prediction method; Sec.~\ref{sec:exp_results} shows the experimental results; and, Sec.~\ref{sec:conclusions} concludes the paper.

\section{Sensor Network}  

\begin{figure}[ht]
\centering
\includegraphics[width=0.3\textwidth]{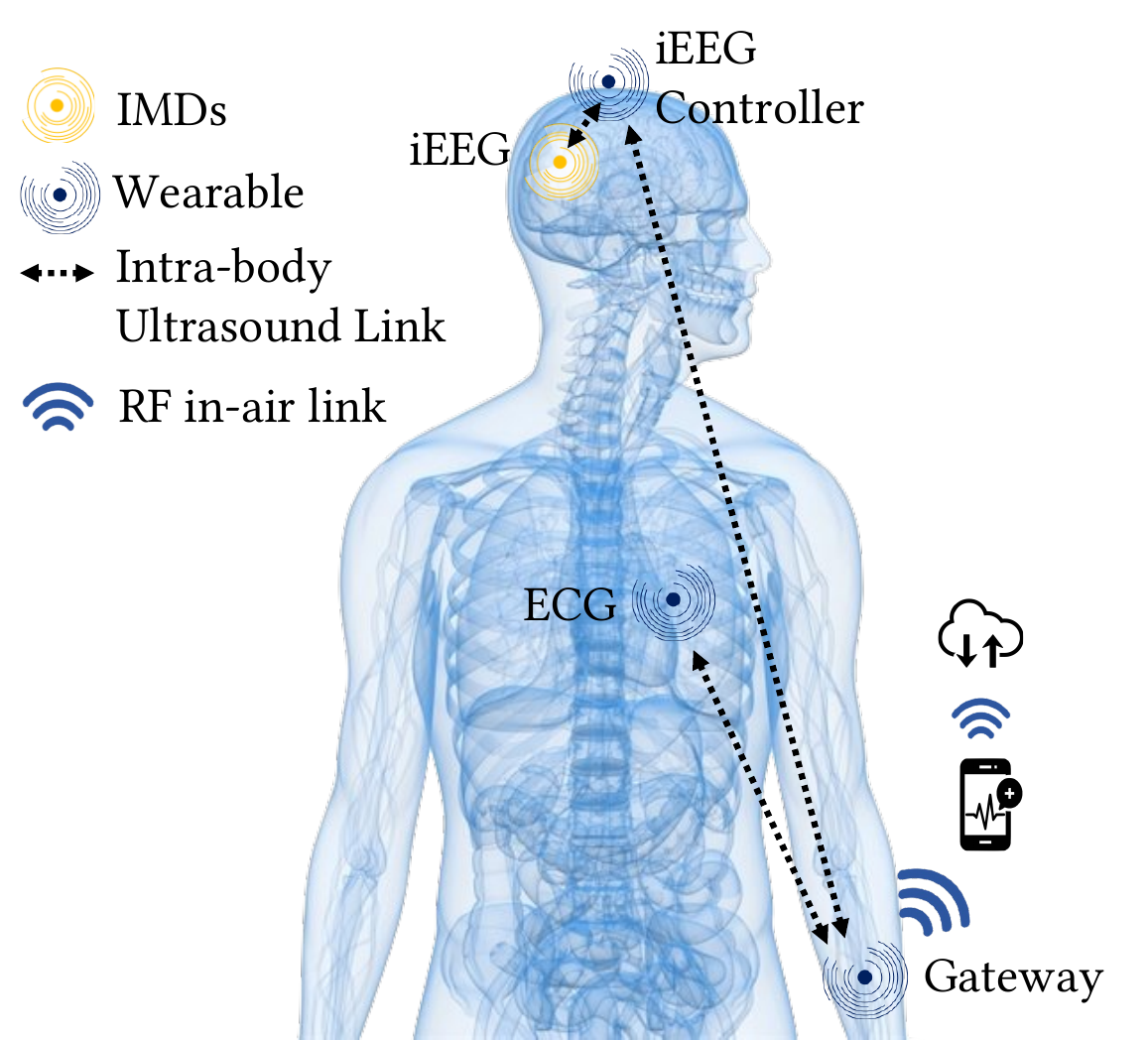}
\caption{System architecture; Gateway receives the classification results from the \gls{ieeg} and \gls{ecg} nodes that execute \gls{dl} algorithms.}
\vspace{-15pt}
\label{fig:system}
\end{figure}

At its core, our proposed system, SeizNet, consists of three wearable or implantable nodes: (i) the \gls{ieeg} classifier; (ii) the \gls{ecg} classifier (both use a \gls{dl} model to process the sensor data); and (iii) the gateway, that receives and combines the \gls{dl} classification results from the two classifiers to make decisions as explained in Sec.~\ref{sec:dl-system}. The nodes in wireless sensor network use an ultrasonic communication platform ~\cite{SantagatiTran20} (see Fig.~\ref{fig:system}).

\label{sec:sensor-network}


Based on the result of the prediction, either an alerting signal can be sent to the patient, or a stimulation command can be sent to a \gls{dbs} system to responsively stimulate targeted areas in the brain and alleviate the effect or prevent a seizure onset from happening \cite{morrell2011responsive}.
This ensures timely intervention and appropriate medical attention.

\noindent\textbf{Physical Layer.}
Short-range and long-range through-tissue ultrasonic communication channel are considered to transmit and receive data between the implanted and wearable nodes that executes \gls{ieeg} and \gls{ecg} data classifications and the wearable gateway.
The short-range (few mm) ultrasonic link allows for focused propagation with lower transmission power that saves energy on the implant. This is a critical aspect for the implant, since its charging operations are more complicated than the wearable nodes, and the energy buffer size has to be small to be easily implanted~\cite{Santagati17INFOCOM}.
Besides the ultrasonic communication transceiver, the gateway uses traditional RF-based communications, such as Bluetooth or Wi-Fi, to exchange data with an Internet enabled device~\cite{SantagatiTran20}.





\noindent\textbf{\gls{mac} Layer.}
The following \emph{application messages} (not including control frames) are exchanged between the sensor nodes and the gateway:
(i) classification results from \gls{ieeg} classifier to \gls{ieeg} controller; (ii) classification results from over-the-skin nodes (\gls{ieeg} controller and \gls{ecg} sensors) to the gateway;
(iii) stimulation settings from the gateway back to the \gls{ieeg} controller (in case of using a closed-loop \gls{dbs} system); and (iv) alert messages from the gateway to an internet-connected devices. 
An impulse based transmission, i.e., a pulse position modulation (PPM), with a superimposed spreading code is used as explained in~\cite{santagati2014medium}.


The outcomes of the \gls{dl} models are encoded into $B_{app}$ bits (application bits) and transmitted every $t_{app}$ seconds. Consequently, the minimum required bit rate $R_{app}$ for each node is calculated as $R_{app} = B_{app} / t_{app}$ in bits per second ($bit/s$). Considering the existence of four nodes (\gls{ecg}, \gls{ieeg}, gateway, and \gls{dbs}), the total bit rate $R_{total}$ equals to $4 \times R_{app}$. The time resolution of the system is $4 s$, implying $t_{app} = 4 s$ for all nodes. Unlike prior works such as \cite{santagati2014medium}, we can adopt a simplified centralized \gls{mac} mechanism. This centralized \gls{mac} protocol is chosen due to the system's inherent characteristics: a fixed number of nodes and a gateway. 
Initially, the gateway coordinates other nodes by dispatching a control message encompassing the spreading code and time-hopping frame sequence assigned to each sensor node.

The employed spreading code and time-hopping frame sequence enable multiple nodes to effectively share the channel, enabling simultaneous communication. This obviates the necessity for control messages to synchronize and mutually exclude nodes, a challenging task in ultrasonic communications due to extended and unpredictable propagation delays. As described in \cite{santagati2014medium}, the \gls{mac} protocol can proficiently support all four nodes, achieving an average throughput of $25 kbit/s$ with a close to $0.005$ packet drop rate. Consequently, each node is allocated $R_{app} = \frac{R_{total}}{N_{nodes}} = 5 kbit/s$. Given that we transmit binary classification results and operate within a time resolution of $4 s$, this bandwidth allocation is more than adequate for the system's requirements.

\section{Dataset} \label{sec:dataset}
We utilize a robust dataset generated under the EPILEPSIAE project \cite{ihle2012epilepsiae}, an EU endeavor, featuring \gls{eeg}, \gls{ieeg} and \gls{ecg} data from 275 focal epilepsy patients. Recorded between 2009 and 2012 at three reputable European Centers, the dataset is known for continuous long-term recordings with an average duration of 165 hours, and and an average of 9.8 seizures per patient. Ultimately, we used 27 patients that have both \gls{ecg} and \gls{ieeg} together. The number and placement of the leads for \gls{ieeg} are different among the patients; however, all of them have single channel \gls{ecg} recorded from their chests. It is worth noting that the \gls{ieeg} signal can be captured with commercial \gls{dbs} systems and used in our proposed solution.

Our objective is to predict seizures an hour in advance of the seizure onset. This is achieved by classifying/differentiating between the pre-seizure and the non-seizure data samples. As seen in Table \ref{tab:seizure_data_distribution}, the vast majority of the data is non-seizure \gls{ieeg} data, amounting to 1.8 billion seconds of data (note that every four seconds of the recording corresponds to one sample in our database). In our analysis, we observed that
the mean ratio of pre-seizure to non-seizure states was about \(0.0826\), with a variance of \(0.0039\). These statistics highlight the imbalance in the dataset, emphasizing the rarity the pre-seizure states compared to non-seizure states.

The data is stored in a postgreSQL database with a relational structure, containing tables for raw \gls{ieeg} and \gls{ecg} data, time references, and a treasure trove of metadata. The metadata encompasses elements such as electrode positions, seizure annotations, medication dosages, patient history, and imaging data, while also containing raw electrode data in binary files.

\begin{table}[!hb]
    \centering
    \begin{tabular}{cccc}
         \toprule
Seizure State & \gls{ecg}      & \gls{eeg}   & \gls{ieeg}       \\
\midrule
Non-Seizure   & 32,016,786 & 465,407,245 & 1,805,183,428 \\
Pre-Seizure   & 2,500,364  & 28,659,972 & 160,301,747   \\
\bottomrule
    \end{tabular}
    \vspace{5pt}
    \caption{Distribution of data sample size for different seizure states}
    \vspace{-15pt}
\label{tab:seizure_data_distribution}
\end{table}


\section{Combined \gls{dl}-Based Seizure Prediction} \label{sec:dl-system}
                                         
\noindent\textbf{Pre-Processing.}
For simplicity and power consumption, high-end pre-processing has been avoided for both \gls{ecg} and \gls{ieeg}. However, the performance of the model has been investigated which was showing that the use of pre-processing including notch-filter for power-line noise and band pass filter does not help improving the results significantly.

\subsection{Deep Learning Model}
\begin{figure}
    \centering
    \includegraphics[width=0.48\textwidth]{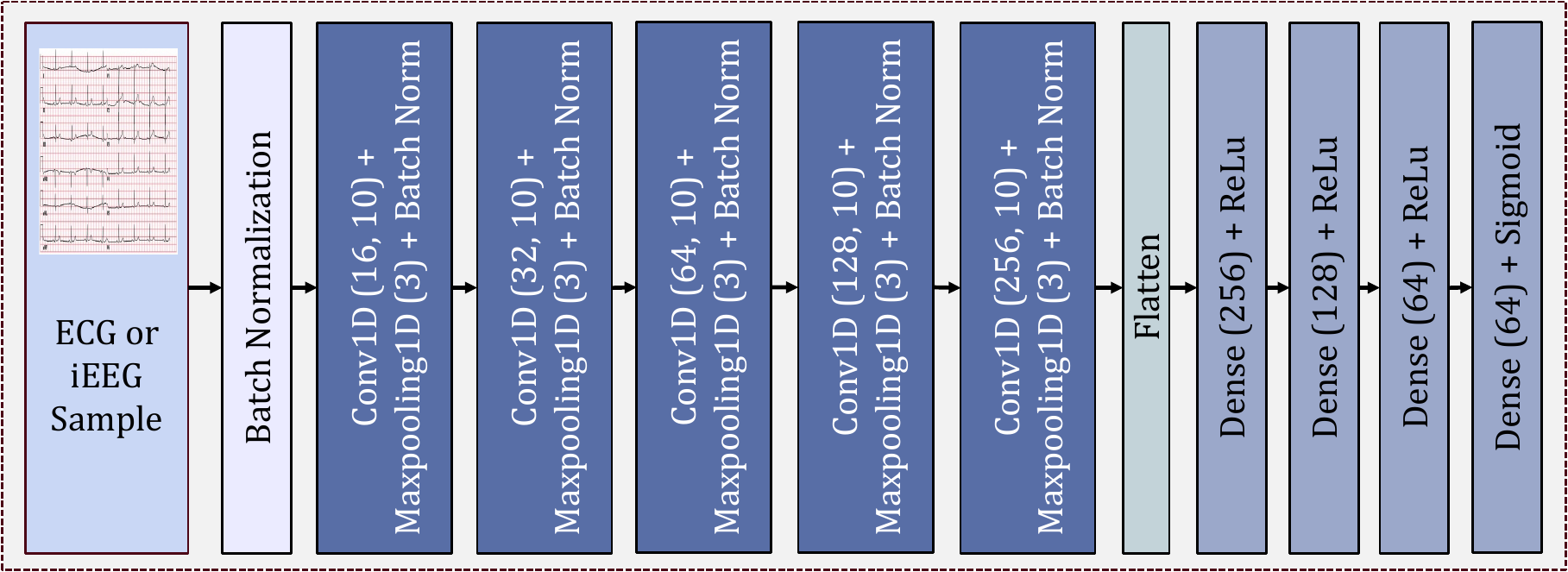}
    \vspace{-7pt}
    \caption{Deep Learning Model Structure.}
    \vspace{-15pt}
    \label{fig:dl_model}
\end{figure}

Our proposed \gls{dl} model for seizure prediction based on \gls{ecg}/\gls{ieeg} recordings consists of multiple stages (Fig.~\ref{fig:dl_model}). Initially, the raw \gls{ecg}/\gls{ieeg} samples undergo batch normalization to enhance their suitability for subsequent processing. These normalized samples then pass through a series of five 1-dimensional \gls{cnn} blocks each accompanying by a Max-pooling layer, in order to extract latent features with maximum information content. The utilization of 1-D Convolutional layers enables the model to effectively capture crucial features from each sample. The resulting features are flattened and subsequently fed into four dense layers for binary classification. The three intermediate layers employ the ReLu, while the final layer utilizes the Sigmoid activation function. We optimized this model to strike a balance between computational efficiency and accuracy.

In our extensive study, we observe that \gls{cnn} models 
are better capable of capturing spatial patterns within \gls{ieeg} and \gls{ecg} signals, extracting nuance features, hence identifying intricate seizure-related patterns more effectively. The \gls{cnn} architecture shown in Fig.~\ref{fig:dl_model} remains consistent across all patients; however, individualized training occurs for each patient. This process results in the creation of a distinct trained model specific to each patient. We allocate 80\% of the dataset for training, 10\% for validation, and 10\% for testing purposes.

\subsection{Focal Loss Function}\label{sec:focal}
Training neural networks for biomedical tasks can be challenging, primarily due to the uneven and inconsistent distribution of labels. This issue persists in the EPILEPSIAE dataset~\cite{ihle2012epilepsiae}, as demonstrated by the fact that the pre-seizure to non-seizure period sample ratio for patients ranges from 0.020 to 0.233. In order to achieve optimal training results, it is crucial to address the class imbalance and leverage all the valuable information contained within the data. To tackle this problem, we employ a novel loss function called Focal loss function \cite{lin2017focal}, which specifically addresses the issue of class imbalance better than balanced cross entropy (BCE) loss function. The focal loss function can be defined as:
\begin{equation}
    FL(p_t) = 
    \begin{cases}
        - \alpha (1 - p)^{\gamma} \log{p}, & y = 1 \\
        - (1 - \alpha) p^{\gamma} \log{1 - p}, & \textit{otherwise}
    \end{cases}
    \label{eqn:focal}
\end{equation}
\noindent where, $p \in [0, 1]$ represents the model's estimated probability for each class and $y$ is the actual label of the class. We consider $y = 1$ for pre-seizure periods and $y = 0$ for non-seizures.

There are two knobs to tune the loss function: $\alpha$ which can be used similar to imbalanced BCE loss function, that puts predefined weight on different classes' loss; and $\gamma$ which helps to improve the behavior of the cross entropy by assigning a lower loss to the misclassified samples.
Through exhaustive search, we identified that setting $\alpha = 0.2$ and $\gamma = 2$ yields the optimal performance across all patients.

This weighting strategy ensures that the model does not favor non-seizure instances, effectively combating the bias and improving the overall classification performance.

\subsection{Time and Channel Voting}
To ensure optimal performance and mitigate the risk of false detection based on a single faulty sample, a majority voting strategy is employed for both channels and time in our proposed approach. For \gls{ecg} signals, which consist of a single channel, time voting is performed by buffering the decisions of the last 15 samples (60 second of the recording) and determining the final decision based on the majority vote. In the case of \gls{ieeg} signals, multiple channels are available, decisions are collected based on buffering both channel and time buffering.

\section{Experimental Results} \label{sec:exp_results}



\subsection{Performance Metrics}
We employ several performance metrics to analyze the SeizNet performance. These metrics include sensitivity, specificity, and accuracy, providing insights into the model's ability to detect pre-seizure periods, non-seizure periods, and overall performance, respectively, as follows: 
\begin{equation}
\text{sensitivity} = \frac{\text{TP}}{\text{TP + FN}}~,~~~ \text{specificity} = \frac{\text{TN}}{\text{TN + FP}}
\end{equation}
\begin{equation}
\text{accuracy} = \frac{\text{TN + TP}}{\text{TN + TP + FP + FN}}
\end{equation}
In addition, we calculate the false positive rate per hour as: 
\begin{equation}
\text{FPR} (h^{-1}) = \frac{\text{FP}}{\text{TN + FP}} \times \frac{3600}{4}
\end{equation}

\subsection{\gls{ecg}-based perdicter with new Focal Loss Function}
\begin{figure}
    \centering
    \includegraphics[width=0.5\textwidth]{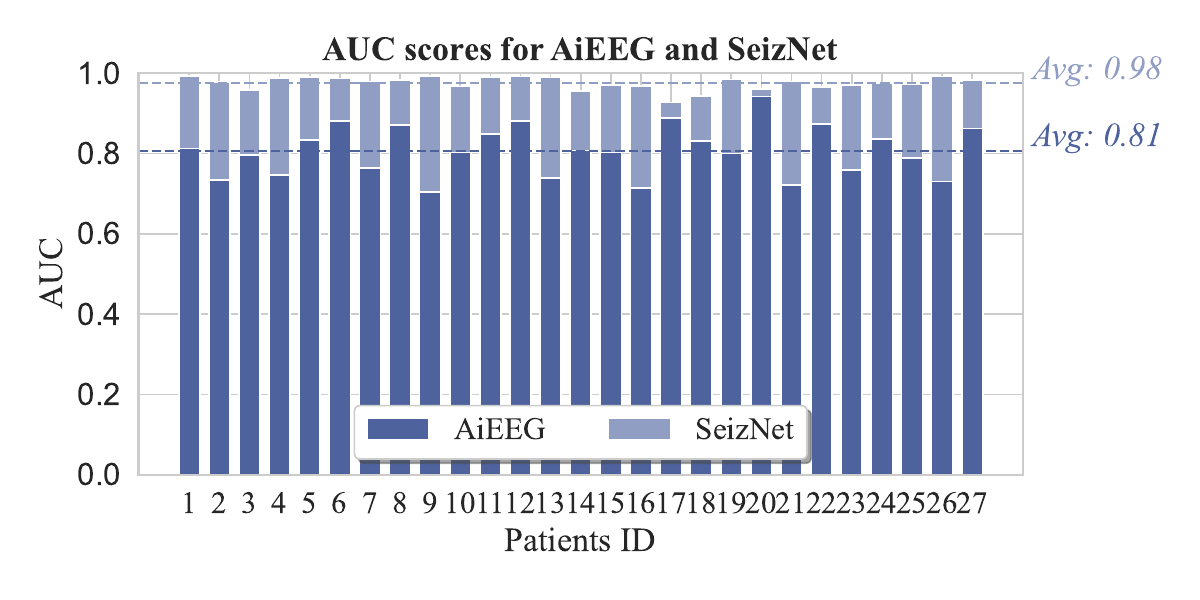}
    \vspace{-15pt}
    \caption{SeizNet \gls{auc} improvements using the new loss function, compared to the baseline model (AiEEG).}
    \vspace{-15pt}
    \label{fig:auc_comparison}
\end{figure}

Fig~\ref{fig:auc_comparison} shows the improvement in \gls{auc} using SeizNet with the proposed focal loss function (Sec.~\ref{sec:focal}) and utilizing \gls{ecg} signal, compared to the \gls{ieeg}-based predictor using the \gls{dl} network structure in AiEEG~\cite{uvaydov2022aieeg}, showing an average 17\% increase (from 81\% to 98\%).

As it can be seen in figure \ref{fig:all-metrics}, thank to the new \gls{dl} structure, the SeizNet network is able to reach up to 99\% accuracy in predicting seizure by only using the non-invasive \gls{ecg} signal, exceeding the performance of the stat-of-the-art~\cite{uvaydov2022aieeg}. Note, that all the seizure prediction are performed up to one hour in advance of a seizure onset.
Leveraging the accessibility of \gls{ecg} signals and their compatibility with smartwatches, this method offers a convenient and practical solution for seizure prediction. However, it is worth noting that the model experiences a relatively high \gls{fph} (up to 4.13 \gls{fph}).
Repetitive false alarms may result in ignoring the true positives, or unneeded stimulation in a closed-loop system with \gls{dbs}.
Addressing this false positive issue will be an essential focus for further refinement and improvement of \gls{ecg} predictors.

\subsection{\gls{ecg}, \gls{ieeg}, and Combined Seizure Predictors}
The \gls{ieeg}-based predictor utilizes multiple channels and adopts time and channel voting for making the final decision (for details on the voting mechanism we refer the readers to~\cite{uvaydov2022aieeg}). As a result, it demonstrates superior performance in avoiding false predictions (achieving accuracy of $99.9\%$) compared to the \gls{ecg} predictor, thanks to its diverse dataset.

\begin{figure}[!b]
    \centering
    \includegraphics[width=0.5\textwidth]{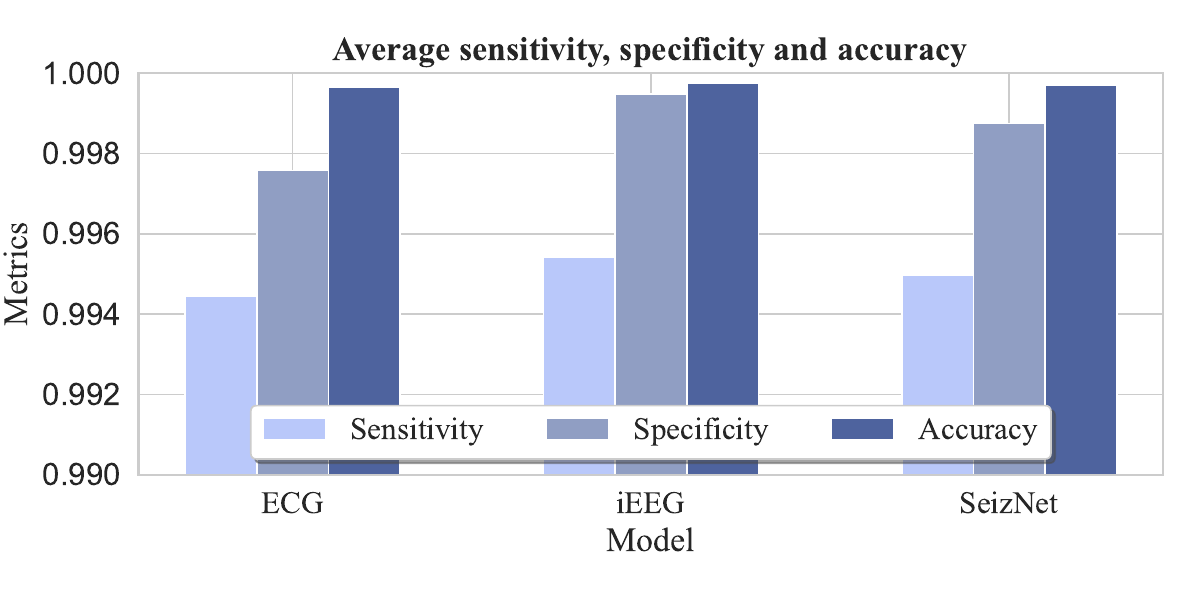}
    \vspace{-15pt}
    \caption{Average Sensitivity, Specificity, and Accuracy among all the patients with \gls{ecg}, \gls{ieeg}, and combined model on test dataset.}
    \vspace{-15pt}
    \label{fig:all-metrics}
\end{figure}


As it can be seen in Fig.~\ref{fig:all-metrics}, the results of both the \gls{ecg} and \gls{ieeg} predictors individually are $>99.9\%$ for all metrics thanks to the new focal loss function and the voting mechanism. 
However, by aggregating the \gls{ecg} and \gls{ieeg} classification results, SeizNet can maintain high accuracy while ensuring minimal false positive rate (as low as 0.23 \gls{fph} thanks to the high sensitivity and specificity). In previous studies, such as \cite{tsiouris2018long}, the most notable performance reached approximately 99\% of sensitivity within a 60-minute pre-ictal time window. Although our results appear comparable, it's important to note that our proposed model, stands out due to its high specificity (very low \gls{fph}) and low computational cost (low weight \gls{cnn} model, without need for additional feature extraction), making it a more practical and feasible method for implementation on resource-restricted medical devices.

\section{Conclusion}\label{sec:conclusions}
In this paper we introduced SeizNet, an AI-enabled sensor network system for seizure prediction, based on \gls{dl} models and advanced data curation techniques. The SeizNet architecture incorporates iEEG and ECG classifiers, connected to a gateway for real-time decision-making. We leveraged a large-scale dataset from the EPILEPSIAE project, and addressed class imbalance challenges through a focal loss function in the proposed \gls{dl} model. The proposed sensor network, comprising implantable and wearable nodes, can form a closed-loop system for effective monitoring, prediction and intervention (as needed) of seizure occurrence. Experimental results show the system's high sensitivity, specificity, and accuracy ($>$99\%) in predicting pre-seizure periods. 

\bibliographystyle{IEEEtran}
\bibliography{Ref}

\end{document}